\documentclass[10pt,twocolumn,letterpaper]{article}
\usepackage{iccv}
\usepackage{times}
\usepackage{epsfig}
\usepackage{graphicx}
\usepackage{amsmath}
\usepackage{amssymb}
\usepackage{multirow}

\usepackage[breaklinks=true,bookmarks=false]{hyperref}

\iccvfinalcopy 

\def\iccvPaperID{467} 

\ificcvfinal\fi
\begin{document}

\title{Unlabeled Samples Generated by GAN \\ Improve the Person Re-identification Baseline \emph{in vitro}}

\author{Zhedong Zheng\qquad Liang Zheng\qquad Yi Yang \thanks{To whom all correspondence should be addressed.} \\
Centre for Artificial Intelligence, University of Technology Sydney\\ 
{\tt\small \{zdzheng12,liangzheng06,yee.i.yang\}@gmail.com}
}

\maketitle
\thispagestyle{empty}

\begin{abstract}
The main contribution of this paper is a simple semi-supervised pipeline that only uses the original training set without collecting extra data. It is challenging in 1) how to obtain more training data only from the training set and 2) how to use the newly generated data. In this work, the generative adversarial network (GAN) is used to generate unlabeled samples. We propose the label smoothing regularization for outliers (LSRO). This method assigns a uniform label distribution to the unlabeled images, which regularizes the supervised model and improves the baseline.

We verify the proposed method on a practical problem: person re-identification (re-ID). This task aims to retrieve a query person from other cameras. We adopt the deep convolutional generative adversarial network (DCGAN) for sample generation, and a baseline convolutional neural network (CNN) for representation learning. Experiments show that adding the GAN-generated data effectively improves the discriminative ability of learned CNN embeddings. On three large-scale datasets, Market-1501, CUHK03 and DukeMTMC-reID, we obtain +4.37\%, +1.6\% and +2.46\% improvement in rank-1 precision over the baseline CNN, respectively. We additionally apply the proposed method to fine-grained bird recognition and achieve a +0.6\% improvement over a strong baseline. The code is available at \url{https://github.com/layumi/Person-reID_GAN}.
\end{abstract}

\section{Introduction}
Unsupervised learning can serve as an important auxiliary task to supervised tasks \cite{hinton2006reducing,rasmus2015semi,goodfellow2013multi,ranzato2008semi}. In this work, we propose a semi-supervised pipeline that works on the original training set without an additional data collection process. First, the training set is expanded with unlabeled data using a GAN. Then our model minimizes the sum of the supervised and the unsupervised losses through a new regularization method. This method is evaluated with person re-ID, which aims to spot the target person in different cameras. This has been recently viewed as an image retrieval problem \cite{zheng2016survey}. 

\begin{figure}[t]
\begin{center}
   \includegraphics[width=1\linewidth]{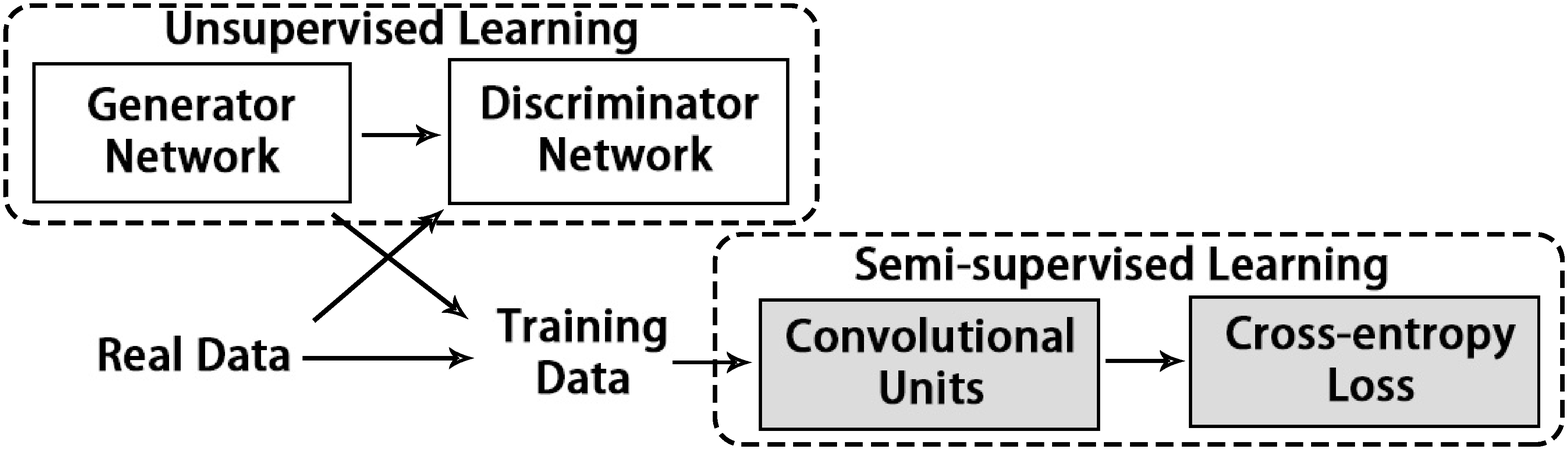}
\end{center}
   \caption{The pipeline of the proposed method. There are two components: a generative adversarial model \cite{radford2015unsupervised} for unsupervised learning and a convolutional neural network for semi-supervised learning. ``Real Data'' represents the labeled data in the given training set; ``Training data'' includes both the ``Real Data'' and the generated unlabeled data. We aim to learn more discriminative embeddings with the ``Training data''. }
\label{fig:pipeline}
\end{figure}

This paper addresses three challenges. First, current research in GANs typically considers the quality of the sample generation with and without semi-supervised learning \emph{in vivo} \cite{odena2016semi,salimans2016improved,radford2015unsupervised,chen2016infogan,pathak2016context,wu2016learning}. Yet a scientific problem remains unknown: moving the generated samples out of the box and using them in currently available learning frameworks. To this end, this work uses unlabeled data produced by the DCGAN model \cite{radford2015unsupervised} in conjunction with the labeled training data. As shown in Fig. \ref{fig:pipeline}, our pipeline feeds the newly generated samples into another learning machine (i.e. a CNN). Therefore, we use the term ``\emph{in vitro}'' to differentiate our method from \cite{odena2016semi,salimans2016improved,radford2015unsupervised,chen2016infogan}; these methods perform semi-supervised learning in the discriminator of the GANs (\emph{in vivo}).

Second, the challenge of performing semi-supervised learning using labeled and unlabeled data in CNN-based methods remains. Usually, the unsupervised data is used as a pre-training step before supervised learning \cite{ranzato2008semi,goodfellow2013multi,hinton2006reducing}. Our method uses all the data simultaneously. In \cite{papandreou2015weakly,lee2013pseudo,odena2016semi,salimans2016improved}, the unlabeled/weak-labeled real data are assigned labels according to pre-defined training classes, but our method assumes that the GAN generated data does not belong to any of the existing classes. The proposed LSRO method neither includes unsupervised pre-training nor label assignments for the known classes. We address semi-supervised learning from a new perspective. Since the unlabeled samples do not belong to any of the existing classes, they are assigned a uniform label distribution over the training classes. The network is trained not to predict a particular class for the generated data with high confidence.

\begin{figure}[t]
\begin{center}
   \includegraphics[width=1.0\linewidth]{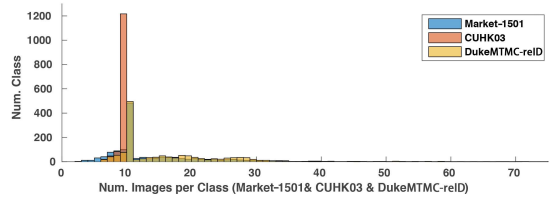}
\end{center}
\caption{The image distribution per class in the dataset Market-1501 \cite{zheng2015scalable}, CUHK03 \cite{li2014deepreid} and DukeMTMC-reID \cite{ristani2016MTMC}. We observe that all these datasets suffer from the limited images per class. Note that there are only a few classes with more than 20 images.}
\label{fig:class}
\end{figure}

Third, in person re-ID, data annotation is expensive, because one has to draw a pedestrian bounding box and assign an ID label to it. Recent progress in this field can be attributed to two factors: 1) the availability of large-scale re-ID datasets \cite{zheng2015scalable,zheng2016person,xiao2016end,li2014deepreid} and 2) the learned embedding of pedestrians using a CNN \cite{cheng2016person,geng2016deep}. That being said, the number of images for each identity is still limited, as shown in Fig. \ref{fig:class}. There are 17.2 images per identities in Market-1501 \cite{zheng2015scalable}, 9.6 images in CUHK03 \cite{li2014deepreid}, and 23.5 images in DukeMTMC-reID \cite{ristani2016MTMC} on average. So using additional data is non-trivial to avoid model overfitting. In the literature, pedestrian images used in training are usually provided by the training sets, without being expanded. So it is unknown if a larger training set with unlabeled images would bring any extra benefit. This observation inspired us to resort to the GAN samples to enlarge and enrich the training set. It also motivated us to employ the proposed regularization to implement a semi-supervised system. 


In an attempt to overcome the above-mentioned challenges, this paper 1) adopts GAN in unlabeled data generation, 2) proposes the label smoothing regularization for outliers (LSRO) for unlabeled data integration, and 3) reports improvements over a CNN baseline on three person re-ID datasets. In more details, in the first step, we train DCGAN \cite{radford2015unsupervised} on the original re-ID training set. We generate new pedestrian images by inputting 100-dim random vectors in which each entry falls within [-1, 1]. Some generated samples are shown in Fig. \ref{fig:imperfect} and Fig. \ref{fig:gan}.  In the second step, these unlabeled GAN-generated data are fed into the ResNet model \cite{he2016deep}. The LSRO method regularizes the learning process by integrating the unlabeled data and, thus, reduces the risk of over-fitting. 
Finally, we evaluate the proposed method on person re-ID and show that the learned embeddings demonstrate a consistent improvement over the strong ResNet baseline.

To summarize, our contributions are:

\begin{itemize}
\item the introduction of a semi-supervised pipeline that integrates GAN-generated images into the CNN learning machine \emph{in vitro};
\item an LSRO method for semi-supervised learning. The integration of unlabeled data regularizes the CNN learning process. We show that the LSRO method is superior to the two available strategies for dealing with unlabeled data; and
\item a demonstration that the proposed semi-supervised pipeline has a consistent improvement over the ResNet baseline on three person re-ID datasets and one fine-grained recognition dataset. 
\end{itemize}

\section{Related Work}
In this section, we will discuss the relevant works on GANs, semi-supervised learning and person re-ID.

\subsection{Generative Adversarial Networks}
The generative adversarial networks (GANs) learn two sub-networks: a generator and a discriminator. The discriminator reveals whether a sample is generated or real, while the generator produces samples to cheat the discriminator. The GANs are first proposed by Goodfellow \etal \cite{goodfellow2014generative} to generate images and gain insights into neural networks. Then, DCGANs \cite{radford2015unsupervised} provides some techniques to improve the stability of training. The discriminator of DCGAN can serve as a robust feature extractor. Salimans \etal \cite{salimans2016improved} achieve a state-of-art result in semi-supervised classification and improves the visual quality of GANs. InfoGAN \cite{chen2016infogan} learns interpretable representations by introducing latent codes. On the other hand, GANs also demonstrate potential in generating images for specific fields. Pathak \etal \cite{pathak2016context} propose an encoder-decoder method for image inpainting, where GANs are used as the image generator. Similarly, Yeh \etal \cite{yeh2016semantic} improve the inpainting performance by introducing two loss types. In \cite{wu2016learning}, 3D object images are generated by a 3D-GAN. In this work, we do not focus on investigating more sophisticated sample generation methods. Instead, we use a basic GAN model \cite{radford2015unsupervised} to generate unlabeled samples from the training data and show that these samples help improve discriminative learning.

\subsection{Semi-supervised Learning}
Semi-supervised learning is a sub-class of supervised learning taking unlabeled data into consideration, especially when the volume of annotated data is small.  On the one hand, some research treats unsupervised learning as an auxiliary task to supervised learning. For example, in \cite{hinton2006reducing}, Hinton \etal learn a stack of unsupervised restricted Boltzmann machines to pre-train the model. Ranzato \etal propose to reconstruct the input at every level of a network to get a compact representation \cite{ranzato2008semi}. In \cite{rasmus2015semi}, the auxiliary task of ladder networks is to denoise representations at every level of the model. On the other hand, several works assign labels to the unlabeled data. Papandreou \etal \cite{papandreou2015weakly} combine strong and weak labels in CNNs using an expectation-maximization (EM) process for image segmentation. In \cite{lee2013pseudo}, Lee assigns a ``pseudo label'' to the unlabeled data in the class that has the maximum predicted probability. In \cite{odena2016semi,salimans2016improved}, the samples produced by the generator of the GAN are all taken as one class in the discriminator. Departing from previous semi-supervised works, we adopt a different regularization approach by assigning a uniform label distribution to the generated samples.

\begin{figure}[t]
\begin{center}
   \includegraphics[width=1\linewidth]{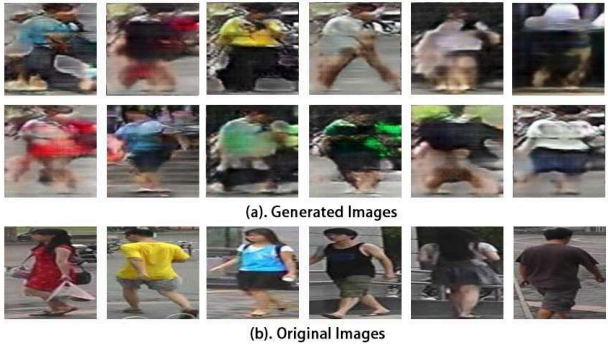}
\end{center}
   \caption{Examples of GAN images and real images. (a) The top two rows show the pedestrian samples generated by DCGAN \cite{radford2015unsupervised} trained on the Market-1501 training set \cite{zheng2015scalable}. (b) The bottom row shows the real samples in training set. Although the generated images in (a) can be easily recognized as fake images by a human, they still serve as an effective regularizer in our experiment. }
\label{fig:imperfect}
\end{figure} 

\subsection{Person Re-identification}
Some pioneering works focus on finding discriminative handcrafted features \cite{ma2012bicov,ma2014covariance,liao2015person}. Recent progress in person re-ID mainly consists of advancing CNNs. Yi \etal \cite{yi2014deep} split a pedestrian image into three horizontal parts and respectively train three part-CNNs to extract features. Similarly, Cheng \etal \cite{cheng2016person} split the convolutional map into four parts and fuse the part features with the global feature. In \cite{li2014deepreid}, Li \etal add a new layer that multiplies the activation of two images in different horizontal stripes. They use this layer to explicitly allow patch matching in the CNN. Later, Ahmed \etal \cite{ahmed2015improved} improve the performance by proposing a new patch matching layer that compares the activation of two images in neighboring pixels. In addition, Varior \etal \cite{varior2016gated} combine the CNN with some gate functions, aiming to adaptively focus on the salient parts of input image pairs, this method is limited by computational inefficiency because the input should be image pairs. 

A CNN can be very discriminative by itself without explicit part-matching. Zheng \etal \cite{zheng2016survey,zheng2016person} directly use a conventional fine-tuning approach (called the ID-discriminative embedding, or IDE) on the Market-1501 dataset \cite{zheng2015scalable} and its performance exceeds many other recent results. Wu \etal \cite{wu2016enhanced} combine the CNN embedding with hand-crafted features. In \cite{zheng2016discriminatively}, Zheng \etal combine an identification model with a verification model and improve the fine-tuned CNN performance. In this work, we adopt the IDE model \cite{zheng2016survey,zheng2016person} as a baseline, and show that the GAN samples and LSRO effectively improve its performance. Recently, Barbosa \etal \cite{barbosa2017looking} propose synthesizing human images through a photorealistic body generation software. These images are used to pre-train an IDE model before dataset-specific fine-tuning. Our method is different from \cite{barbosa2017looking} in both data generation and the training strategy.

\section{Network Overview}
In this section, we describe the pipeline of the proposed method. As shown in Fig. \ref{fig:pipeline}, the real data in the training set is used to train the GAN model. Then, the real training data and the newly generated samples are combined into training input for the CNN. In the following section, we will illustrate the structure of the two components, \ie, the GAN and the CNN, in detail. Note that, \textbf{our system does not make major changes to the network structures of the GAN or the CNN with one exception - the number of neurons in the last fully-connected layer in the CNN is modified according to the number of training classes.}

\subsection{Generative Adversarial Network}
Generative adversarial networks have two components: a generator and a discriminator. For the generator, we follow the settings in \cite{radford2015unsupervised}. We start with a 100-dim random vector and enlarge it to $4\times4\times16$ using a linear function. To enlarge the tensor, five deconvolution functions are used with a kernel size of $5\times5$ and a stride of $2$. Every deconvolution is followed by a rectified linear unit and batch normalization. Additionally, one optional deconvolutional layer with a kernel size of $5\times5$ and a stride of $1$, and one \emph{tanh} function are added to fine-tune the result. A sample that is $128\times128\times3$ in size can then be generated. 
 
The input of the discriminator network includes the generated images and the real images in the training set. We use five convolutional layers to classify whether the generated image is fake. Similarly, the size of the convolutional filters is $5\times5$ and their stride is $2$. We add a fully-connected layer to perform the binary classification (real or fake).

\subsection{Convolutional Neural Network}
The ResNet-50 \cite{he2016deep} model is used in our experiment. We resize the generated images to $256\times256\times3$ using bilinear sampling. The generated images are mixed with the original training set as the input of the CNN. That is, the labeled and unlabeled data are simultaneously trained. These training images are shuffled. Following the conventional fine-tuning strategy \cite{zheng2016survey}, we use a model pre-trained on ImageNet \cite{russakovsky2015imagenet}. We modify the last fully-connected layer to have $K$ neurons to predict the $K$-classes, where $K$ is the number of the classes in the original training set (as well as the merged new training set). Unlike \cite{odena2016semi,salimans2016improved}, we do not view the new samples as an extra class but assign a uniform label distribution over the existing classes. So the last fully-connected layer remains $K$-dimensional. The assigned label distribution of the generated images is discussed in the next section.

\section{The Proposed Regularization Method} \label{loss}
In this section, we first revisit the label smoothing regularization (LSR), which is used for fully-supervised learning. We then extend LSR to the scenario of unlabeled learning, yielding the proposed label smoothing regularization for outliers (LSRO) method. 


\subsection{Label Smoothing Regularization Revisit}\label{sec:revisit}
LSR was proposed in the 1980s and recently re-discovered by Szegedy \etal \cite{szegedy2016rethinking}. In a nutshell, LSR assigns small values to the non-ground truth classes instead of $0$. This strategy discourages the network to be tuned towards the ground truth class and thus reduces the chances of over-fitting. LSR is proposed for use with the cross-entropy loss \cite{szegedy2016rethinking}. 

Formally, let $k\in\{1,2,...,K\}$ be the pre-defined classes of the training data, where $K$ is the number of classes. The cross-entropy loss can be formulated as:
\begin{equation}
l = -\sum_{k=1}^{K} \log{(p(k))}q(k),
\label{Eq.1}
\end{equation}
where $p(k)\in[0,1]$ is the predicted probability of the input belonging to class $k$, and can be outputted by CNN. It is derived from the softmax function which normalizes the output of the previous fully-connected layer. $q(k)$ is the ground truth distribution. Let y be the ground truth class label, $q(k)$ can be defined as:
\begin{equation}
    q(k)=
   \begin{cases}
   0 &\mbox{$k \ne y$}\\
   1 &\mbox{$k = y$}
   \end{cases}.
\label{Eq.2}
\end{equation}
If we discard the $0$ terms in Eq. \ref{Eq.1}, the cross-entropy loss is equivalent to only considering the ground truth term in Eq. \ref{Eq.3}. 
\begin{equation}
    l = - \log{(p(y))}.
\label{Eq.3}
\end{equation}
So, minimizing the cross-entropy loss is equivalent to maximizing the predicted probability of the ground-truth class.
In \cite{szegedy2016rethinking}, the label smoothing regularization (LSR) is introduced to take the distribution of the non-ground truth classes into account. The network is thus encouraged not to be too confident towards the ground truth. In \cite{szegedy2016rethinking}, the label distribution $q_{LSR}(k)$ is written as:
\begin{equation}
    q_{LSR}(k)=
   \begin{cases}
   \frac{\varepsilon}{K} &\mbox{ $k\ne y$ }\\
   1-\varepsilon+\frac{\varepsilon}{K} &\mbox{ $k=y$ }
   \end{cases},
\label{Eq.4}
\end{equation}
where $\varepsilon \in{[0,1]}$ is a hyperparameter. If $\varepsilon$ is zero, Eq. \ref{Eq.4} reduces to Eq. \ref{Eq.2}. If $\varepsilon$ is too large, the model may fail to predict the ground truth label. So in most cases, $\varepsilon$ is set to $0.1$. 
Szegedy \etal assume that the non-ground truth classes take on a uniform label distribution. Considering Eq. \ref{Eq.1} and Eq. \ref{Eq.4}, the cross-entropy loss evolves to:
\begin{equation}
l_{LSR} = - (1-\varepsilon)\log{(p(y))} - \frac{\varepsilon}{K}\sum_{k=1}^{K} \log{(p(k))}.
\label{Eq.5}
\end{equation}
Compared with Eq. \ref{Eq.3}, Eq. \ref{Eq.5} pays additional attention to the other classes, rather than only the ground truth class. In this paper, we do not employ LSR on the IDE baseline because it yields a slightly lower performance than using Eq. \ref{Eq.2} (see Section \ref{sec:Evaluation}). We re-introduce LSR because it inspires us in designing the LSRO method.

\begin{figure}[t]
\begin{center}
   \includegraphics[width=1\linewidth]{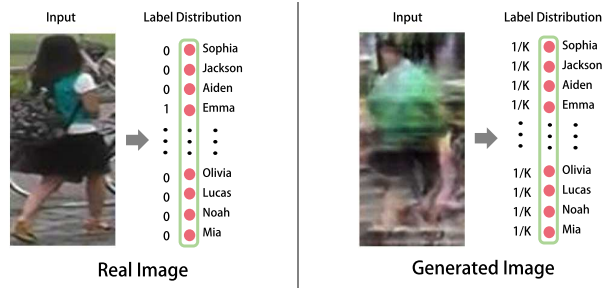}
\end{center}
   \caption{The label distributions of a real image and a GAN-generated image in our system. We use a classical label distribution (Eq. \ref{Eq.2}) for the real image (left). For the generated image (right), we employ the proposed LSRO label distribution (Eq. \ref{Eq.6}), \eg a uniform distribution on every training class because the generated image is assumed to belong to none of the training classes. We employ a cross-entropy loss that combines the two types of label distributions as the optimization objective (Eq. \ref{Eq.7}). }
\label{fig:method}
\end{figure} 

\subsection{Label Smoothing Regularization for Outliers}
The label smoothing regularization for outliers (LSRO) is used to incorporate the unlabeled images in the network. This extends LSR from the supervised domain to leverage unsupervised data generated by the GAN. 

In LSRO, we propose a virtual label distribution for the unlabeled images. We set the virtual label distribution to be uniform over all classes, due to two inspirations. 1) We assume that the generated samples do not belong to any pre-defined classes. 2) LSR assumes a uniform distribution over the all classes to address over-fitting. During testing, we expect that the maximum class probability of a generated image will be low, \ie, the network will fail to predict a particular class with high confidence. Formally, for a generated image, its class label distribution, $q_{LSRO}(k)$, is defined as:
\begin{equation}
    q_{LSRO}(k)= \frac{1}{K}.
\label{Eq.6}
\end{equation}
We call Eq. \ref{Eq.6} the label smoothing regularization for outliers (LSRO). 

The one-hot distribution defined in Eq. \ref{Eq.2} will still be used for the loss computation for the real images in the training set. 
Combining Eq. \ref{Eq.2}, Eq. \ref{Eq.6} and Eq. \ref{Eq.1}, we can re-write the cross-entropy loss as:
\begin{equation}
l_{LSRO} = -(1-Z) \log{(p(y))} 
- \frac{Z}{K} \sum_{k=1}^{K} \log{(p(k))}.
\label{Eq.7}
\end{equation}
For a real training image, $Z=0$. For a generated training image, $Z=1$. So our system actually has two types of losses, one for real images and one for generated images. 


\textbf{Advantage of LSRO.} Using LSRO, we can deal with more training images (outliers) that are located near the real training images in the sample space, and introduce more color, lighting and pose variances to regularize the model. For instance, if we only have one green-clothed identity in the training set, the network may be misled into considering that the color green is a discriminative feature, and this limits the discriminative ability of the model. By adding generated training samples, such as an unlabeled green-clothed person, the classifier will be penalized if it makes the wrong prediction towards the labeled green-clothed person. In this manner, we encourage the network to find more underlying causes and to be less prone to over-fitting. We only use the GAN trained on the original training set to produce outlier images. It would be interesting to further evaluate whether real-world unlabeled images are able to achieve a similar effect (see Table \ref{table:addcuhk}).

\textbf{Competing methods.} We compare LSRO with two alternative methods. Details of both methods are available in existing literature \cite{odena2016semi,salimans2016improved,lee2013pseudo}; breif descriptions follow.
\begin{itemize}
\item \textbf{All in one.} Using \cite{odena2016semi,salimans2016improved}, a new class label is created, \ie, $K+1$, and every generated sample is assigned to this class. CNN training follows in Section \ref{sec:details}.
\item \textbf{Pseudo label.} Using \cite{lee2013pseudo}, during network training, each incoming GAN-image is passed forward through the current network and is assigned a pseudo label by taking the maximum value of the probability prediction vector ($p(k)$ in Eq. \ref{Eq.1}). This GAN-image can be thus trained in the network with this pseudo label. During training, the pseudo label is assigned \emph{dynamically}, so that the same GAN-image may receive different pseudo labels each time it is fed into the network. In our experiments, we begin feeding GAN images and assigning them pseudo labels after 20 epochs. We also set a global weight to the softmax loss of 0.1 to the GAN and 1 to the real images.
\end{itemize}

Our experimental results show that the two methods also work on the GAN images and that LSRO is superior to ``All in one'' and ``Pseudo label''. Explanations are provided in the Section \ref{sec:Evaluation}.




\section{Experiment} \label{sec:experiment}
We mainly evaluate the proposed method using the Market-1501 \cite{zheng2015scalable} dataset, because it is a large scale and has a fixed training/testing split. We also report results on the CUHK03 dataset \cite{li2014deepreid}, \textbf{but due to the computational cost of 20 training/testing splits, we only use the GAN images generated from the Market-1501 dataset}. In addition, we evaluate our method on a recently released pedestrian dataset DukeMTMC-reID \cite{ristani2016MTMC} and a fine-grained recognition dataset CUB-200-2011 \cite{WahCUB_200_2011}. 

\subsection{Person Re-id Datasets}
\textbf{Market-1501} is a large-scale person re-ID dataset collected from six cameras. It contains 19,732 images for testing and 12,936 images for training. The images are automatically detected by the deformable part model (DPM) \cite{felzenszwalb2010object}, so misalignment is common, and the dataset is close to realistic settings. There are 751 identities in the training set and 750 identities in the testing set. There are 17.2 images per identity in the training set. We use all the 12,936 detected images from the training set to train the GAN. 

\textbf{CUHK03} contains 14,097 images of 1,467 identities. Each identity is captured by two cameras on the CUHK campus. This dataset contains two image sets. One is annotated by hand-drawn bounding boxes, and the other is produced by the DPM detector \cite{felzenszwalb2010object}. We use the detected set in this paper. There are 9.6 images per identity in the training set. We report the averaged result after training/testing 20 times. We use the \textbf{single shot} setting.

\begin{figure}
\begin{center}
   \includegraphics[width=1\linewidth]{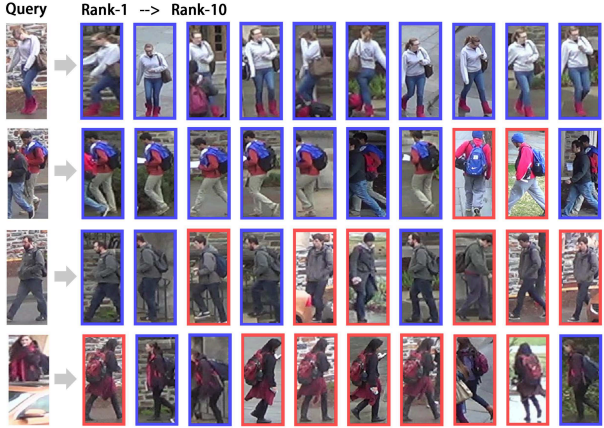}
\end{center}
   \caption{Sample retrieval results on DukeMTMC-reID using the proposed method. The images in the first column are the query images. The retrieved images are sorted according to the similarity scores from left to right. The correct matches are in the blue rectangles, and the false matching images are in the red rectangles. DukeMTMC-reID is challenging because it contains pedestrians with occlusions and similar appearance.}
\label{fig:duke}
\end{figure}

\textbf{DukeMTMC-reID} is a subset of the newly-released multi-target, multi-camera pedestrian tracking dataset \cite{ristani2016MTMC}. The original dataset contains eight 85-minute high-resolution videos from eight different cameras. Hand-drawn pedestrian bounding boxes are available. In this work, we use a subset of \cite{ristani2016MTMC} for image-based re-ID, in the format of the Market-1501 dataset \cite{zheng2015scalable}. We crop pedestrian images from the videos every 120 frames, yielding 36,411 total bounding boxes with IDs annotated by \cite{ristani2016MTMC}. The DukeMTMC-reID dataset for re-ID  has 1,812 identities from eight cameras. There are 1,404 identities appearing in more than two cameras and 408 identities (distractor ID) who appear in only one camera. We randomly select 702 IDs as the training set and the remaining 702 IDs as the testing set. In the testing set, we pick one query image for each ID in each camera and put the remaining images in the gallery. As a result, we get 16,522 training images with 702 identities, 2,228 query images of the other 702 identities and 17,661 gallery images. The evaluation protocol is available on our website \cite{DukeMTMC-reID}. Some example re-ID results from the DukeMTMC-reID are shown in Fig. \ref{fig:duke}.

\subsection{Implementation Details}\label{sec:details}
\textbf{CNN re-ID baseline.} We adopt the CNN re-ID baseline used in \cite{zheng2016survey,zheng2016person}. Specifically, the Matconvnet \cite{vedaldi15matconvnet} package is used. During training, We use the ResNet-50 model \cite{he2016deep} and modify the fully-connected layer to have 751, 702 and 1,367 neurons for Market-1501, DukeMTMC-reID and CUHK03, respectively. All the images are resized to $256\times256$ before being randomly cropped into $224\times224$ with random horizontal flipping. We insert a dropout layer before the final convolutional layer and set the dropout rate to 0.5 for CUHK03 and 0.75 for Market-1501 and DukeMTMC-reID, respectively. We use stochastic gradient descent with momentum 0.9. The learning rate of the convolution layers is set to 0.002 and decay to 0.0002 after 40 epochs and we stop training after the 50th epochs. During testing, we extract the 2,048-dim CNN embedding in the last convolutional layer for an $224\times224$ input image. The similarity between two images is calculated by a cosine distance for ranking.

\begin{figure}
\begin{center}
   \includegraphics[width=1\linewidth]{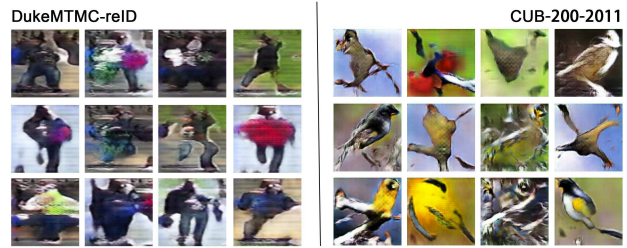}
\end{center}
   \caption{The newly generated images from a DCGAN model trained on DukeMTMC-reID and CUB-200-2011. Through LSRO, they are added to the training sets of DukeMTMC-reID and CUB-200-2011 to regularize the CNN model.}
\label{fig:gan}
\end{figure}

\textbf{GAN training and testing.} We use Tensorflow \cite{abadi2016tensorflow} and the DCGAN package \cite{DCGAN-tensorflow} to train the GAN model using the provided data in the original training set without preprocessing (\eg, foreground detection). All the images are resized to $128\times128$ and randomly flipped before training. We use Adam \cite{kingma2014adam} with the parameters $\beta_1=0.5, \beta_2=0.99$. We stop training after 30 epochs. During GAN testing, we input a 100-dim random vector in GAN, and the value of each entry ranges in [-1, 1]. The outputted image is resized to $256\times256$ and then used in CNN training (with LSRO). More GAN images are shown in Fig. \ref{fig:gan}.


\subsection{Evaluation} \label{sec:Evaluation}
\textbf{The ResNet baseline.} Using the training/testing procedure described in Section \ref{sec:details}, we report the baseline performance of ResNet in Table \ref{table:mr}, Table \ref{table:CUHK03} and Table \ref{table:duke}. The rank-1 accuracy is 73.69\%, 71.5\% and 60.28\% on Market-1501, CUHK03 and DukeMTMC-reID respectively. Our baseline results are on par with the those reported in \cite{zheng2016survey,zheng2016discriminatively}. Note that the baseline alone exceeds many previous works \cite{liao2015person,varior2016siamese,zhang2016learning}.


\textbf{The GAN images improve the baseline.}
As shown in Table \ref{table:bigdata}, when we add $24,000$ GAN images to the CNN training, our method significantly improves the re-ID performance on Market-1501. We observe improvement of +4.37\% (from 73.69\% to 78.06\%) and +4.75\% (from 51.48\% to 56.23\%) in rank-1 accuracy and mAP, respectively. On CUHK03, we observe improvements of +1.6\%, +1.2\%, +0.8\%, and +1.6\% in rank-1, 5, 10 accuracy and mAP, respectively. The improvement on CUHK03 is relatively small compared to that of Market-1501, because the DCGAN model is trained on Market-1501 and the generated images share a more similar distribution with Market-1501 than CUHK03. We also observe improvements of +2.46\% and +2.14\% in rank-1 and mAP, respectively, on the strong ResNet baseline in the DukeMTMC-reID dataset. These results indicate that the unlabeled images generated by the GAN effectively yield improvements over the baseline using the LSRO method. 
\setlength{\tabcolsep}{5.7pt}
\begin{table}
\begin{center}
\begin{tabular}{l|cc|cc}
\hline
\multirow{2}{*}{method} & \multicolumn{2}{c|}{Single Query} & \multicolumn{2}{c}{Multi. Query}\\
& rank-1 & mAP & rank-1  & mAP \\
\hline
BoW+kissme \cite{zheng2015scalable} & 44.42 & 20.76  & - & -\\
MR CNN \cite{ustinova2015multiregion} & 45.58 & 26.11 & 56.59 & 32.26 \\
FisherNet \cite{wu2016deep} & 48.15 & 29.94 & - & -\\
SL \cite{chen2016similarity} & 51.90 & 26.35  & - & -\\
S-LSTM \cite{varior2016siamese} & - & - & 61.6 & 35.3 \\
DNS \cite{zhang2016learning} & 55.43 & 29.87 & 71.56 & 46.03 \\
Gate Reid \cite{varior2016gated}  & 65.88 & 39.55& 76.04 & 48.45 \\
SOMAnet \cite{barbosa2017looking}*& 73.87 & 47.89 & 81.29 & 56.98 \\ 
Verif.-Identif. \cite{zheng2016discriminatively}* & 79.51 & 59.87  & 85.84 & 70.33 \\
DeepTransfer \cite{geng2016deep}* & 83.7 & 65.5 & \textbf{89.6} & 73.8 \\ 
\hline
Basel. \cite{zheng2016survey,zheng2016discriminatively}* & 73.69 & 51.48 & 81.47 & 63.95 \\
Basel. + LSRO & 78.06 & 56.23  & 85.12 & 68.52 \\
Verif-Identif. + LSRO & \textbf{83.97} & \textbf{66.07} & 88.42 & \textbf{76.10}  \\
\hline
\end{tabular}
\end{center}
\caption{Comparison of the state-of-the-art methods reported on the Market-1501 dataset. We also provide results of the fine-tuned ResNet baseline. Rank-1 precision (\%) and mAP (\%) are listed. * the respective paper is on ArXiv but not published.}
\label{table:mr}
\end{table}

\setlength{\tabcolsep}{4pt}
\begin{table}
\small
\begin{center}
\begin{tabular}{l|cc|cc|cc}
\hline
\multirow{2}{*}{\# GAN Img.} & \multicolumn{2}{c|}{LSRO} & \multicolumn{2}{c|}{All in one} & \multicolumn{2}{c}{Pseudo label}\\
& rank-1 & mAP & rank-1  & mAP& rank-1  & mAP \\
\hline
0 (basel.) & 73.69 & 51.48  & 73.69 & 51.48  & 73.69 & 51.48\\ 
12,000 & 76.81 & 55.32 & 75.33 & 52.82 & 76.07 & \textcolor{red}{53.56} \\    
18,000 & 77.26 & 55.55 & \textcolor{blue}{77.20} & 55.04 & \textcolor{red}{76.34} & 53.45\\
24,000 & \textbf{78.06} & \textbf{56.23} & 76.63 & 55.12 & 75.80 & 53.03\\
30,000 & 77.38 & 55.48& 75.95 &55.18& 75.21 & 52.65 \\ 
36,000 & 76.07 & 54.59& 76.87 &\textcolor{blue}{55.47} & 74.67 & 52.38\\
\hline
\end{tabular}
\end{center}
\caption{Comparison of LSRO, ``All in one'', and ``Pseudo label'' under different numbers of GAN-generated images on Market-1501. We show that LSRO is superior to the other two methods whose best performance is highlighted in \textcolor{blue}{blue} and \textcolor{red}{red}, respectively. Rank-1 accuracy (\%) and mAP (\%) are shown.}
\label{table:bigdata}
\end{table}

\textbf{The impact of using different numbers of GAN images during training.} 
We evaluate how the number of GAN images affects the re-ID performance. Since unlabeled data is easy to obtain, we expect the model would learn more general knowledge as the number of unlabeled images increases. The results on Market-1501 are shown in Table \ref{table:bigdata}. We note that the number of real training images in Market-1501 is 12,936. Two observations are made. 

First, the addition of different numbers of GAN images consistently improves the baseline. Adding approximately 3$\times$GAN images compared to the real training set still has a +2.38\% improvement to rank-1 accuracy. 

Second, the peak performance is achieved when 2$\times$GAN images are added. When too few GAN sample are incorporated into the system, the regularization ability of the LSRO is inadequate. In contrast, when too many GAN samples are present, the learning machine tends to converge towards assigning uniform prediction probabilities to all the training samples, which is not desirable. Therefore, a trade-off is recommended to avoid poor regularization and over-fitting of uniform label distributions.

\setlength{\tabcolsep}{15pt}
\begin{table}
\begin{center}
\begin{tabular}{l|cc}
\hline
method & rank-1 & mAP \\
\hline
BoW+kissme \cite{zheng2015scalable} & 25.13 & 12.17 \\
LOMO+XQDA \cite{liao2015person} & 30.75 & 17.04\\
\hline
Basel. \cite{zheng2016survey,zheng2016discriminatively} & 65.22 & 44.99\\ 
Basel. + LSRO & \textbf{67.68}  & \textbf{47.13}\\
\hline
\end{tabular}
\end{center}
\caption{Comparison of the baseline on DukeMTMC-reID. Rank-1 accuracy (\%) and mAP (\%) are shown.}
\label{table:duke}
\end{table}

\setlength{\tabcolsep}{10pt}
\begin{table}
\begin{center}
\begin{tabular}{l|cc}
\hline
Unsup. Data & rank-1 & mAP \\
\hline
0 (basel.) & 73.69 & 51.48\\ 
CUHK03-Real-12000 & 75.65  & 53.25 \\
Market-1501-GAN-12000 & \textbf{76.81} & \textbf{55.32} \\
\hline
\end{tabular}
\end{center}
\caption{We add the 12,000 real pedestrian images in CUHK03 as outliers to Market-1501. We find the model trained on the generated samples slightly out-performs the model trained on CUHK03 real data. Rank-1 accuracy (\%) and mAP (\%) are shown.}
\label{table:addcuhk}
\end{table}


\textbf{GAN images vs. real images in training.} To further evaluate the proposed method, we replace the GAN images with the real images from CUHK03 which are viewed as unlabeled in training. Since CUHK03 only contains 14,097 images, we randomly select 12,000 for the fair comparison. 

Experimental results are shown in Table \ref{table:addcuhk}. We compare the results obtained using the 12,000 CUHK03 images and the 12,000 GAN images. We find the real data from CUHK03 also assists in the regularization and improves the performance. But the model trained with GAN-generated data is sightly better. In fact, although the images generated from DCGAN are visually imperfect (see Fig. \ref{fig:imperfect}), they still possess similar regularization ability as the real images.

\setlength{\tabcolsep}{4pt}
\begin{table}
\begin{center}
\begin{tabular}{l|ccccc}
\hline
method & rank-1 & rank-5 & rank-10 & mAP\\
\hline
KISSME \cite{kostinger2012large} & 11.7 & 33.3 & 48.0 & -\\
DeepReID \cite{li2014deepreid} & 19.9 & 49.3 & 64.7 & -\\ 
BoW+HS \cite{zheng2015scalable} & 24.3 & - & - & -\\
LOMO+XQDA \cite{liao2015person} & 46.3 & 78.9 & 88.6 &-\\
SI-CI \cite{wang2016joint} & 52.2 & 84.3 & 94.8 &-\\
DNS \cite{zhang2016learning} & 54.7 & 80.1 & 88.3 &-\\
SOMAnet \cite{barbosa2017looking}* &72.4& 92.1 & 95.8 & - \\
Verif-Identif. \cite{zheng2016discriminatively}* & 83.4 & 97.1 & 98.7 & 86.4\\
DeepTransfer \cite{geng2016deep}* & 84.1& - & - & - \\
\hline
Basel. \cite{zheng2016survey,zheng2016discriminatively}* & 71.5 & 91.5 & 95.9 & 75.8\\ 
Basel.+LSRO & 73.1 & 92.7 & 96.7 & 77.4 \\
Verif-Identif. + LSRO & \textbf{84.6} & \textbf{97.6} & \textbf{98.9} & \textbf{87.4}\\
\hline
\end{tabular}
\end{center}
\caption{Comparison of the state-of-the-art reports on the CUHK03 dataset. We list the fine-tuned ResNet baseline as well. The mAP (\%) and rank1 (\%) precision are presented. * the respective paper is on ArXiv but not published.}
\label{table:CUHK03}
\end{table}

\textbf{Comparison with the two competing methods.}
We compare the LSRO method with the ``All in one'' and ``Pseudo label'' methods implied in \cite{odena2016semi,salimans2016improved} and \cite{lee2013pseudo}, respectively. The experimental results on Market-1501 are summarized in Table \ref{table:bigdata}. 

We first observe that both strategies yield improvement over the baseline. The ``All in one'' method treats all the unlabeled samples as a new class, which forces the network to make ``careful'' predictions for the existing $K$ classes. The ``Pseudo label'' method gradually labels the new data, and thus introduces more variance to the network. 

Nevertheless, we find that LSRO exceeds both strategies by approximately +1\% $\sim$ +2\%. We speculate the reason is that the ``All in one'' method makes a coarse label estimation, while the ``Pseudo label'' originally assumes that all the unlabeled data belongs to the existing classes \cite{lee2013pseudo} which is not true in person re-ID. While these two methods still use the one-hot label distribution, the LSRO method makes a less stronger assumption (label smoothing) towards the labels of the GAN images. These reasons may explain why LSRO has a superior performance.

\textbf{Comparison with the state-of-the-art methods.} 
We compare our method with the state-of-the-art methods on Market-1501 and CUHK03, listed in Table \ref{table:mr} and Table \ref{table:CUHK03}, respectively. On Market-1501, we achieve \textbf{rank-1 accuracy = 78.06\%, mAP = 56.23\%} when using the single query mode, which is the best result compared to the published papers, and the second best among all the available results including ArXiv papers. On CUHK03, we arrive at \textbf{rank-1 accuracy = 73.1\%, mAP = 77.4\%} which is also very competitive. The previous best result is produced by combining the identification and verification losses \cite{geng2016deep,zheng2016discriminatively}. We further investigate whether the LSRO could work on this model. We fine-tuned the publicly available model in \cite{zheng2016discriminatively} with LSRO and achieve state-of-the-art results \textbf{rank-1 accuracy = 83.97\%, mAP = 66.07\%} on Market-1501. On CUHK03, we also observe a state-of-the art performance \textbf{rank-1 accuracy = 84.6\%, mAP = 87.4\%}. We, therefore, show that the LSRO method is complementary to previous methods due to the regularization of the GAN data. 



\setlength{\tabcolsep}{6pt}
\begin{table}
\begin{center}
\begin{tabular}{l|ccc}
\hline
method & model & annotation & top-1\\
\hline
Zhang \etal \cite{zhang2014part}& AlexNet & 2$\times$part & 76.7 \\
Zhang \etal \cite{zhang2014part}& VGGNet & 2$\times$part & 81.6 \\
Liu \etal \cite{liu2016localizing} & ResNet-50 & attribute & 82.9\\
Wang \etal \cite{wang2015multiple} & 3$\times$VGGNet & $\times$ & 83.0\\
\hline
Basel. \cite{liu2016localizing} & ResNet-50 & $\times$ & 82.6\\
Basel.+LSRO & ResNet-50 & $\times$ & 83.2 \\
Basel.+LSRO & 2$\times$ResNet-50 & $\times$ & \textbf{84.4} \\
\hline
\end{tabular}
\end{center}
\caption{We show the recognition accuracy (\%) on CUB-200-2011. The proposed method has a 0.6\% improvement over the competitive baseline. The two-model ensemble shows a competitive result.}
\label{table:cub}
\end{table}

\subsection{Fine-grained Recognition}
Fine-grained recognition also faces the problem of a lack of training data and annotations. To further test the effectiveness of our method, we provide results on the CUB-200-2011 dataset \cite{WahCUB_200_2011}. This dataset contains 200 bird classes with 29.97 training images per class on average. Bounding boxes are used in both training and testing. We do not use part annotations.
In our implementation, the ResNet baseline has a recognition accuracy of 82.6\%, which is slightly higher than the 82.3\% reported in \cite{liu2016localizing}. This is the baseline we will compare our method with.

Using the same pipeline in Fig. \ref{fig:pipeline}, we train DCGAN on the 5,994 images in the training set, and then we combine the real images with the generated images (see Fig. \ref{fig:gan}) to train the CNN. During testing, we adopt the standard 10-crop testing \cite{krizhevsky2012imagenet}, which uses $256\times256$ images as input and the averaged prediction as the classification result. As shown in Table \ref{table:cub}, the strong baseline outperforms some recent methods, and the proposed method further yields an improvement of +0.6\% (from 82.6\% to 83.2\%).
We also combine the two models generated by our method with different initializations to form an ensemble. This leads to a \textbf{84.4\%} recognition accuracy. In \cite{liu2016localizing}, Liu \etal report a 85.5\% accuracy with a five-model ensemble using parts and a global scene. We do not include this result because extra annotations are used. We focus on the regularization ability of the GAN, but not on producing a state-of-the-art result.

\section{Conclusion}
In this paper, we propose an ``\emph{in vitro}'' usage of the GANs for representation learning, \ie, person re-identification. Using a baseline DCGAN model \cite{radford2015unsupervised}, we show that the imperfect GAN images effectively demonstrate their regularization ability when trained with a ResNet baseline model. Through the proposed LSRO method, we mix the unlabeled GAN images with the labeled real training images for simultaneous semi-supervised learning. Albeit simple, we demonstrate consistent performance improvement over the re-ID and fine-grained recognition baseline systems, which sheds light on the practical use of GAN-generated data. 

In the future, we will continue to investigate on whether GAN images of better visual quality yield superior results when integrated into supervised learning. This paper provides some baseline evaluations using the imperfect GAN images and the future investigation would be intriguing. 

\textbf{Acknowledgements.} We thank the support of Data to Decisions Cooperative Research Centre (\url{www.d2dcrc.com.au}), Google Faculty Research Award and NVIDIA Corporation with the donation of TITAN X (Pascal) GPU.

{
\footnotesize
\bibliographystyle{ieee}
\bibliography{egbib}
}

\end{document}